\documentclass[11pt,a4paper]{article}
\usepackage[hyperref]{acl2021}
\usepackage{times}
\usepackage{latexsym}

\usepackage{microtype}
\usepackage{linguex}
\usepackage{url}
\usepackage{caption,subcaption}
\usepackage{epsfig}

\aclfinalcopy %
\def\aclpaperid{2889} %

\title{Uncovering Constraint-Based Behavior in Neural Models via Targeted Fine-Tuning}

\author{Forrest Davis \and Marten van Schijndel \\
        Department of Linguistics\\
        Cornell University\\
        \texttt{\{fd252|mv443\}@cornell.edu}}

\date{}

\begin{document}
\maketitle
\begin{abstract}

A growing body of literature has focused on detailing the linguistic knowledge embedded in large, pretrained language models. Existing work has shown that non-linguistic biases in models can drive model behavior away from linguistic generalizations. We hypothesized that competing linguistic processes within a language, rather than just non-linguistic model biases, could obscure underlying linguistic knowledge. We tested this claim by exploring a single phenomenon in four languages: English, Chinese, Spanish, and Italian. While human behavior has been found to be similar across languages, we find cross-linguistic variation in model behavior. We show that competing processes in a language act as constraints on model behavior and demonstrate that targeted fine-tuning can re-weight the learned constraints, uncovering otherwise dormant linguistic knowledge in models. Our results suggest that models need to learn both the linguistic constraints in a language and their relative ranking, with mismatches in either producing non-human-like behavior.

\end{abstract}

\section{Introduction}

Ever larger pretrained language models continue to demonstrate success on a variety of NLP benchmarks \cite[e.g.,][]{ devlinBERTPretrainingDeep2019,brownLanguageModelsAre2020a}.
One common approach for understanding why these models are successful 
is centered on inferring what linguistic knowledge such models 
acquire \cite[e.g.,][]{linzenAssessingAbilityLSTMs2016,hewittStructuralProbeFinding2019,huSystematicAssessmentSyntactic2020,warstadtBLiMPBenchmarkLinguistic2020}. 
Linguistic knowledge alone, of course, does not fully account for model behavior; 
non-linguistic heuristics have also been shown to drive model behavior \cite[e.g., sentence length; see][]{mccoyRightWrongReasons2019,warstadtLearningWhichFeatures2020}.
Nevertheless, when looking across a variety of experimental methods, models appear to acquire some grammatical knowledge 
\cite[see][]{warstadtInvestigatingBERTKnowledge2019}.

However, investigations of linguistic knowledge in language models 
are limited by the overwhelming prominence of work solely on English
\cite[though see][]{gulordavaColorlessGreenRecurrent2018,ravfogelCanLSTMLearn2018,muellerCrossLinguisticSyntacticEvaluation2020a}. 
Prior work has shown non-linguistic biases of 
neural language models
mimic English-like
linguistic structure, limiting the generalizability of claims founded on English data
\cite[e.g.,][]{dyerCriticalAnalysisBiased2019,davis-van-schijndel-2020-recurrent}. 
In the present study, we show via cross-linguistic comparison,
that knowledge of competing linguistic constraints can obscure underlying linguistic knowledge. 

Our investigation is centered on a single discourse phenomena, implicit causality (IC) verbs, in 
four languages: English, Chinese, Spanish, and Italian. When an IC verb occurs in a 
sentence, interpretations of pronouns are affected:

\ex. \label{ic_exp}
    \a. Lavender frightened Kate because she was so terrifying. \label{ic_exp_a}
    \b. Lavender admired Kate because she was so amazing. \label{ic_exp_b}

In \ref{ic_exp}, both \textit{Lavender} and \textit{Kate} agree in gender with \textit{she}, so 
both are possible antecedents. However, English speakers overwhelmingly 
interpret \textit{she} as referring to \textit{Lavender} in \ref{ic_exp_a} and \textit{Kate} 
in \ref{ic_exp_b}. Verbs that 
have a subject preference (e.g., \textit{frightened}) are called subject-biased IC verbs, and
verbs with an object preference (e.g., \textit{admired}) are called object-biased IC verbs.

IC has been a rich source of psycholinguistic investigation \cite[e.g.,][]{garveyImplicitCausalityVerbs1974, hartshorneWhatImplicitCausality2014, williamsLanguageExperiencePredicts2020}.
Current accounts of IC ground the phenomenon within the linguistic signal without the
need for additional pragmatic inferences by comprehenders \cite[e.g.,][]{rohdeAnticipatingExplanationsRelative2011, hartshorneAreImplicitCausality2013}. Recent investigations of IC in neural language models confirms 
that the IC bias of English is learnable, at least to some degree, from text data alone
\cite{davis-van-schijndel-2020-discourse,upadhyePredictingReferenceWhat2020}. The 
ability of models trained on other languages to acquire an IC bias, however, has not been explored. Within the psycholinguistic literature, IC has been shown to be 
remarkably consistent cross-linguistically \cite[see][]{hartshorneAreImplicitCausality2013,ngoImplicitCausalityComparison2020}. 
That is, IC verbs have 
been attested in a variety of languages.
Given the cross-linguistic consistency of 
IC, then, models trained on other languages should also demonstrate an IC bias.
However, using two popular model types, 
BERT based \cite{devlinBERTPretrainingDeep2019} and RoBERTa based \cite{liuRoBERTaRobustlyOptimized2019},\footnote{These model types were chosen for ease 
of access to existing models. Pretrained, large auto-regressive models are largely 
restricted to English, and prior work suggests that LSTMs are limited in their ability 
to acquire an IC bias in English \protect\cite{davis-van-schijndel-2020-discourse}.} 
we find that
models only acquired a human-like IC bias in English and Chinese but not in Spanish and Italian.

We relate this 
to a crucial difference in the presence of 
a competing linguistic constraint 
affecting pronouns in the target languages. 
Namely, Spanish and Italian 
have a well studied process called \textit{pro drop}, which
allows for subjects to be `empty' \cite{rizziNullObjectsItalian1986}.
An English equivalent would be ``(she) likes BERT'' where \textit{she} can be elided. 
While IC verbs increase the probability of a pronoun that refers to a particular antecedent,
pro drop disprefers any overt 
pronoun in subject position (i.e.\ the 
target location in our study). That is, 
both processes are in direct competition 
in our experiments. As a result, Spanish and Italian
models are susceptible to overgeneralizing
any learned pro-drop knowledge, favoring 
no pronouns rather than IC-conditioned 
pronoun generation. 

To exhibit an IC bias, models of Spanish and Italian
have two tasks: learn the relevant constraints (i.e.\ IC and pro drop) and
the relative ranking of these constraints. We find that the models learn both constraints, 
but, critically, instantiate the wrong ranking, favoring pro drop to an IC bias. 
Using fine-tuning to demote pro drop, we are able to uncover otherwise dormant IC knowledge 
in Spanish and Italian. Thus, the apparent failure of the Spanish and Italian models to pattern like English and 
Chinese is not evidence on its own
of a model's inability to acquire the requisite linguistic 
knowledge, but is in fact evidence that models are unable to adjudicate between 
competing linguistic constraints in a human-like way. In English and Chinese, the 
promotion of a pro-drop process via fine-tuning has the opposing effect, 
diminishing an IC bias in model behavior. As such, our results 
indicate that non-human like behavior can be driven by failure either to learn the underlying linguistic constraints or to learn the relevant constraint ranking.

\section{Related Work}

This work is intimately related to the growing body of literature investigating 
linguistic knowledge in large, pretrained models. Largely, this literature articulates 
model knowledge via isolated linguistic phenomena, such as subject-verb agreement \cite[e.g.,][]{linzenAssessingAbilityLSTMs2016,muellerCrossLinguisticSyntacticEvaluation2020a}, 
negative polarity items
\cite[e.g.,][]{marvinTargetedSyntacticEvaluation2018,warstadtInvestigatingBERTKnowledge2019},
and discourse and pragmatic structure \cite[including implicit causality; e.g.,][]{ettingerWhatBERTNot2020a, schusterHarnessingLinguisticSignal2020,jereticAreNaturalLanguage2020,upadhyePredictingReferenceWhat2020}. Our study
differs, largely, in framing model linguistic knowledge as sets of competing constraints, which 
privileges the interaction between linguistic phenomena.

Prior work has noted competing generalizations influencing model behavior via 
the distinction of non-linguistic vs.\ linguistic biases \cite[e.g.,][]{mccoyRightWrongReasons2019,davis-van-schijndel-2020-discourse,warstadtLearningWhichFeatures2020}. 
The findings in \citet{warstadtLearningWhichFeatures2020}, that linguistic knowledge is 
represented within a model much earlier than attestation in model behavior, bears resemblance 
to our claims. We find that linguistic knowledge can, in fact, lie dormant due to other 
linguistic processes in a language, not just due to non-linguistic preferences. Our findings suggest that some linguistic knowledge may never surface in model behavior, though further work is needed on this point. 
\begin{table}[]
    \centering
    \begin{tabular}{l|l|c}
         \textbf{Model} & \textbf{Lang} & \textbf{Tokens}  \\
         \hline
         BERT & EN & 3.3B \\
         RoBERTa & EN & 30B \\
         Chinese BERT & ZH & 5.4B \\
         Chinese RoBERTa & ZH  & 5.4B \\
         BETO & ES & 3B \\
         RuPERTa & ES & 3B \\\
         Italian BERT & IT &  2B\\
         UmBERTo & IT & 0.6B \\
         GilBERTo & IT & 11B\\
         
    \end{tabular}
    \caption{Summary of models investigated with
    language and 
    approximate number of tokens in training. For RoBERTa we use 
    the approximation given in \citet{warstadtLearningWhichFeatures2020}.}
    \label{tab:models}
\end{table}

In the construction of our experiments, we were inspired by synthetic language studies which probe the underlying linguistic capabilities of language models \cite[e.g.,][]{mccoyRevisitingPovertyStimulus2018,ravfogelStudyingInductiveBiases2019}. We made use of synthetically modified language data that accentuated, or weakened, evidence for certain linguistic 
processes. The goal of such modification in our work is quite similar both to work which attempts to 
remove targeted linguistic knowledge in model representations \cite[e.g.,][]{ravfogelNullItOut2020,elazarAmnesicProbingBehavioral2021} 
and to work which investigates the representational space of models via priming \cite{prasadUsingPrimingUncover2019,misraExploringBERTSensitivity2020a}. In the 
present study, rather than identifying isolated linguistic knowledge or using priming to study relations between underlying linguistic representations, we ask \textbf{how linguistic representations interact to drive model behavior}. 

\section{Models}

Prior work on IC in neural language models has been restricted to autoregressive models 
for ease of comparison to human results \cite[e.g.,][]{upadhyePredictingReferenceWhat2020}. 
In 
the present study, we focused on two popular non-autoregressive language model variants, BERT 
\cite{devlinBERTPretrainingDeep2019} and RoBERTa \cite{liuRoBERTaRobustlyOptimized2019}.
We used existing models available via HuggingFace \cite{wolfTransformersStateoftheArtNatural2020}. 

Multilingual models 
have been claimed to perform worse on targeted linguistics tasks
than monolingual models \cite[e.g.,][]{muellerCrossLinguisticSyntacticEvaluation2020a}. 
We confirmed this claim by evaluating mBERT 
which exhibited no IC bias in any language.\footnote{Results are provided in Appendix \ref{sec:full_results}}
Thus, we focus in the rest of this paper on 
monolingual models (summarized in Table \ref{tab:models}).
For English, we used the BERT base uncased model and the RoBERTa base model.
For Chinese, we evaluated BERT and RoBERTa models from \citet{cuiRevisitingPreTrainedModels2020}.
For Spanish, we used BETO \cite{CaneteCFP2020} 
and RuPERTa \cite{romeroRuPERTaSpanishRoBERTa2020}.
For Italian, we evaluated an uncased Italian BERT \footnote{https://huggingface.co/dbmdz/bert-base-italian-uncased}
as well as two RoBERTa based models, UmBERTo \cite{parisiUmBERToItalianLanguage2020}
and GilBERTo \cite{ravasioGilBERToItalianPretrained2020}.

\section{Experimental Stimuli and Measures}

Our list of target verbs was derived from existing psycholinguistic studies of IC verbs.\footnote{All stimuli, as well as code for reproducing the results 
of the paper are available at \url{https://github.com/forrestdavis/ImplicitCausality}
. For each language investigated, the stimuli were evaluated for grammaticality by native speakers with academic training in linguistics.} For English, 
we used the IC verbs from \citet{ferstlImplicitCausalityBias2011}. 

Each verb in the human 
experiment was coded for IC bias based on continuations of sentence fragments (e.g., 
\textit{Kate accused Bill because ...}). For Spanish, we used the IC verbs from \citet{goikoetxeaNormativeStudyImplicit2008}, which followed 
a similar paradigm as \citet{ferstlImplicitCausalityBias2011} for English. Participants were given 
sentence fragments and asked to complete the sentence and circle their intended referent. The study 
reported the percent of subject continuations for 100 verbs, from which we used the 61 verbs which had
a significant IC bias (i.e.\ excluding verbs with no significant subject or object bias). 

For Italian, we used the 40 IC verbs reported in \citet{mannettiInterpersonalVerbsImplicit1991a}. Human 
participants were given ambiguous completed sentences with no overt pronoun 
like ``John feared Michael because of the kind of person (he) is'' and were asked to judge who the null 
pronoun referred to, with the average number of responses that gave the subject as the antecedent reported.\footnote{Specifically, 
\citet{mannettiInterpersonalVerbsImplicit1991a} grouped the verbs into four categories and reported the average per category as well 
as individual verb results for the most biased verbs and the negative/positive valency verbs. Additionally, figures showing 
average responses across various conditions was reported for one of the categories. From the combination of this
information, the average scores for all but two verbs were able to be determined. The remaining two verbs were
assigned the reported average score of their stimuli group.} For Chinese, we used 59 IC verbs reported in 
\citet{hartshorneAreImplicitCausality2013}, which determined average subject bias per verb in a similar way as 
\citet{mannettiInterpersonalVerbsImplicit1991a} (i.e.\ judgments of antecedent preferences given ambiguous sentences, 
this time with overt pronouns).\footnote{In \citet{hartshorneAreImplicitCausality2013}, 60 verbs were reported, but after 
consultation with a native speaker with academic training in linguistics, one verb was excluded due to perceived ungrammaticality of the 
construction.}

We generated stimuli using 14 pairs of stereotypical male and female nouns (e.g., \textit{man} vs.\ \textit{woman}, \textit{husband} vs.\ \textit{wife}) in each language, rather than rely on proper names as was done in the human experiments. The models we investigated are bidirectional, so 
we used a neutral right context, \textit{was there}, for English and Spanish, where 
human experiments provided no right context.\footnote{Using \textit{here, outside,} or \textit{inside} as the right context produces qualitatively the same patterns.}
For Italian we utilized the full sentences
investigated in the human experiments. 
The Chinese human experiment also used full sentences, but relied on nonce words (i.e.\ novel, constructed words like sliktopoz), so 
we chose instead to generate sentences
like the English and Spanish ones.
All stimuli had subjects and objects that 
differed in gender, such that all nouns occurred in subject or object position (i.e.\ the stimuli were fully 
balanced for gender):

\ex. \label{bert_stim} the man admired the woman because [MASK] was there.\footnote{The model-specific mask token was used. Additionally, all models were uncased, with the exception of RoBERTa, so lower cased stimuli were used.}

The mismatch in gender 
forced the choice of pronoun to be unambiguous. For each stimulus, we gathered the scores assigned to the 
third person singular male and female pronouns (e.g., \textit{he} and \textit{she}).\footnote{In spoken Chinese, the male and female pronouns are homophonous. They 
are, however, distinguished in writing.}
Our measures were grouped by antecedent type (i.e.\ the pronoun refers to the 
subject or the object) and whether the verb was object-biased or subject-biased. 
For example, BERT assigns to \ref{bert_stim} 
a score of 0.01 for the subject antecedent (i.e.\ \textit{he}) and 0.97 for the object (i.e.\ \textit{she}), in line with the object-bias 
of \textit{admire}.

\begin{figure}[t] 
    \includegraphics[width=\linewidth]{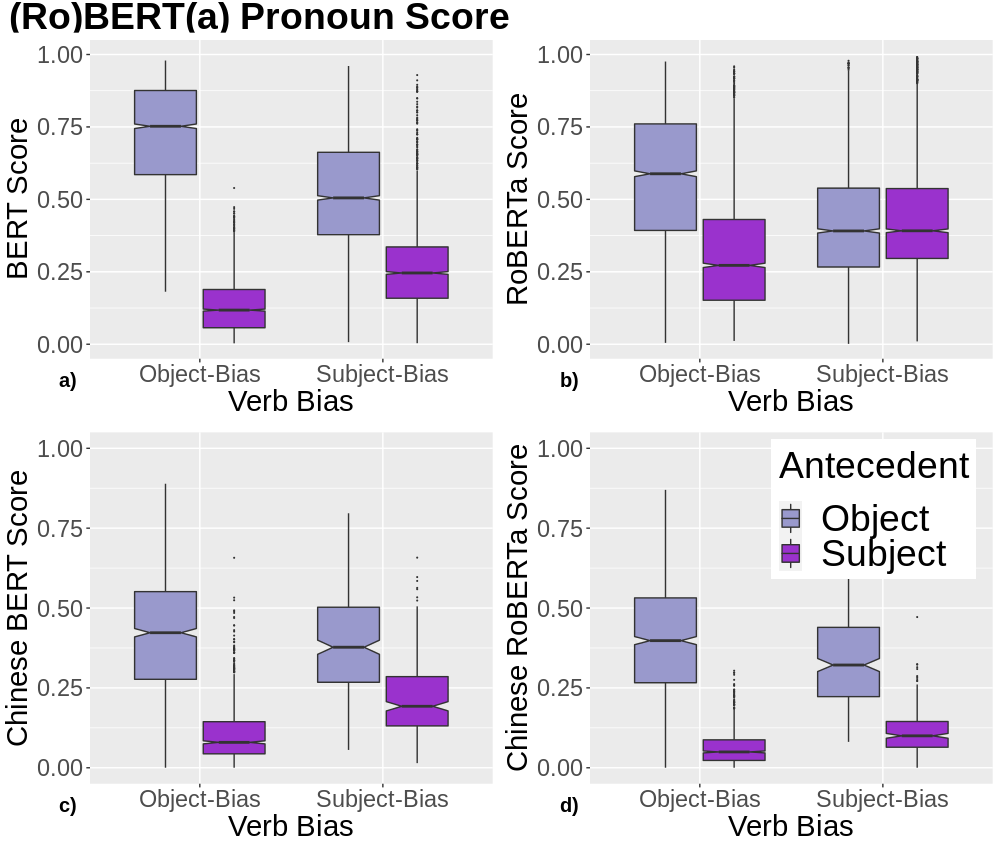}
    \caption{Model scores for \textbf{a)} BERT, \textbf{b)} RoBERTa, \textbf{c)} Chinese BERT, 
    and \textbf{d)} Chinese RoBERTa at the pronoun grouped by antecedent; stimuli
    derived from \citet{ferstlImplicitCausalityBias2011} and \citet{hartshorneAreImplicitCausality2013}}
    \label{fig:en_zh_pronoun_og}
\end{figure}

\section{Models Inconsistently Capture Implicit Causality}

As exemplified in \ref{ic_exp}, repeated below, IC verb bias modulates the preference for pronouns. 

\ex. \label{redux_ic_exp}
    \a. Lavender frightened Kate because she was so terrifying. \label{redux_ic_exp_a}
    \b. Lavender admired Kate because she was so amazing. \label{redux_ic_exp_b}

An object-biased IC verb (e.g., \textit{admired}) should increase the likelihood of pronouns that refer to 
the object, and a subject-biased IC verb (e.g., \textit{frightened}) should increase the likelihood of reference 
to the subject. Given that all the investigated stimuli were disambiguated by gender, 
we categorized our results by the antecedent of the pronoun and the IC verb bias. We 
first turn to English and Chinese, which showed an IC bias in line with existing work 
on IC bias in autoregressive English models \cite[e.g.,][]{upadhyePredictingReferenceWhat2020,davis-van-schijndel-2020-discourse}. 
We then detail the results for Spanish and Italian, where only very limited, if 
any, IC bias was observed.

\begin{figure}[t] 
    \includegraphics[width=\linewidth]{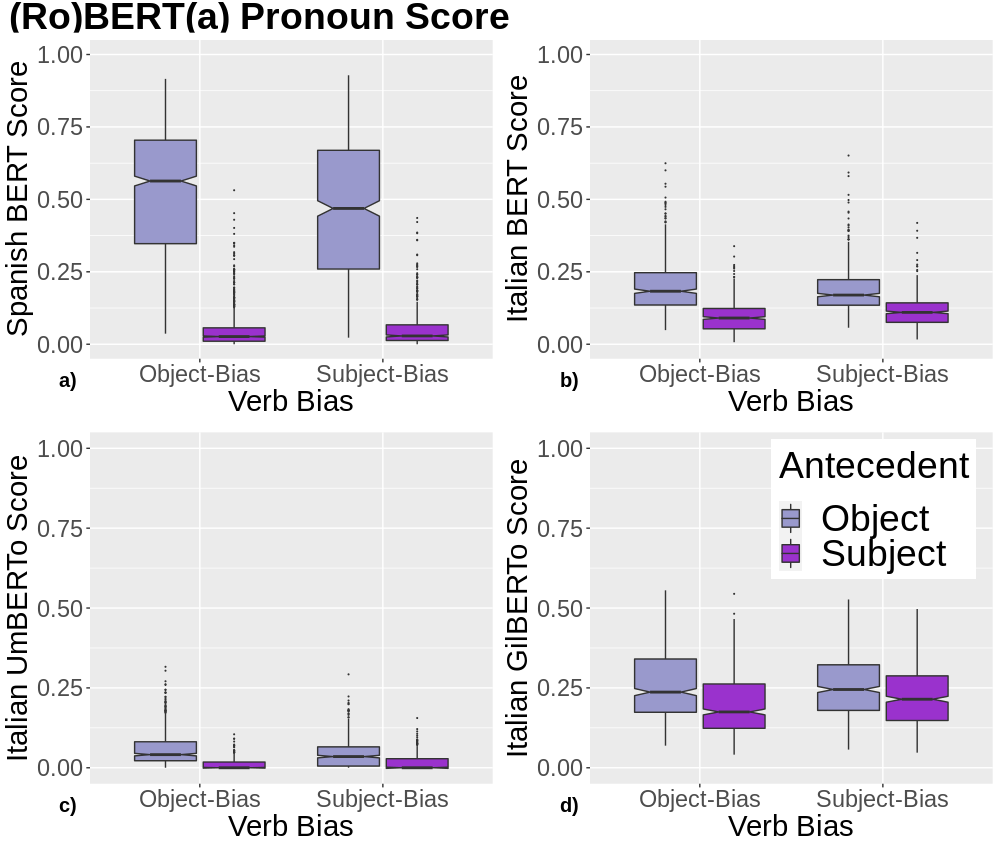}
    \caption{Model scores for \textbf{a)} Spanish BERT (BETO), \textbf{b)} Italian BERT, \textbf{c)} UmBERTo, 
    and \textbf{d)} GilBERTo at the pronoun grouped by antecedent; stimuli
    derived from \citet{goikoetxeaNormativeStudyImplicit2008} and \citet{mannettiInterpersonalVerbsImplicit1991a}}
    \label{fig:es_it_pronoun_og}
\end{figure}

\subsection{English and Chinese} \label{sec:en_zh_og}

The results for English and Chinese are given in Figure \ref{fig:en_zh_pronoun_og} and 
detailed in 
Appendix \ref{sec:full_results}. All models
demonstrated a
greater preference for pronouns referring to the object after an object-biased IC verb
than after a subject-biased IC verb.\footnote{Throughout the paper, statistical significance was determined by two-way \textit{t}-tests evaluating the difference between pronouns referring
to objects after subject-biased and object-biased IC verbs, and similarly for pronouns 
referring to the subject. The threshold for statistical significance was p = 0.0009, after 
adjusting for the 54 statistical tests conducted in the paper.} Additionally, they had greater preferences for 
pronouns referring to the subject after a subject-biased IC verb than after a object-biased 
IC verb. That is, all models showed the
expected IC-bias effect. Generally, there was an overall greater preference for referring to the object, 
in line with a recency bias, with the exception of RoBERTa, where subject-biased IC verbs
neutralized the recency effect. 

\subsection{Spanish and Italian} \label{sec:es_it_og}

The results for Spanish and Italian are given in Figure \ref{fig:es_it_pronoun_og} and 
detailed in Appendix \ref{sec:full_results}. 
In stark contrast to the models of English and Chinese, an IC bias was either not demonstrated 
or was only weakly attested. For Spanish, BETO showed a greater preference for pronouns referencing the object
after an object-biased IC verb than after a subject-biased IC verb. 
There was no 
corresponding IC effect for pronouns referring to the subject, and RuPERTa (a RoBERTa based model)
had no IC effect at all. 

Italian BERT and GilBERTo (a RoBERTa based model)
had no significant effect of IC-verb on pronouns referring to the object. 
There was a significant, albeit very small, %
increased score for 
pronouns referring to the subject after a subject-biased IC verb in line with a weak 
subject-IC bias. Similarly, UmBERTo (a RoBERTa based model) had significant, yet tiny %
IC effects, where object-biased IC verbs increased the score of pronouns referring 
to objects compared to subject-biased IC verbs (conversely with pronouns referring to the 
subject). 

Any significant 
effects in Spanish and Italian were much smaller than their counterparts in English 
(as is visually apparent between Figure \ref{fig:en_zh_pronoun_og} and Figure \ref{fig:es_it_pronoun_og}), and each of the Spanish 
and Italian
models failed to demonstrate at least 
one of the IC effects. 

\section{Pro Drop and Implicit Causality: Competing Constraints}

We were left with an apparent mismatch between models of English and Chinese 
and models of Spanish and Italian. In the former, an IC verb bias modulated
pronoun preferences. In the latter, the same IC verb bias was comparably absent. 
Recall that, for humans, the psycholinguistic literature suggests that IC
bias is, in fact, quite consistent across
languages \cite[see][]{hartshorneAreImplicitCausality2013}. 

We found a possible reason for why the two sets of models behave so differently by carefully considering the languages under investigation. 
Languages can be thought of as systems of competing linguistic constraints \cite[e.g., Optimality Theory;][]{princeOptimalityTheoryConstraint2004}. Spanish and 
Italian exhibit pro drop and
typical grammatical sentences often 
lack overt pronouns in subject position, 
opting instead to rely on rich agreement systems to disambiguate the intended subject
at the verb \cite{rizziNullObjectsItalian1986}. 
This constraint competes with IC, which favors pronouns
that refer to either the subject or the 
object. Chinese also allows for empty arguments (both subjects and objects), typically called \textit{discourse 
pro-drop} \cite{huangDistributionReferenceEmpty1984}.%
\footnote{Other names common to the literature include \textit{topic drop}, 
\textit{radical pro drop}, and \textit{rampant pro drop}.} As the name suggests, however,
this process is more discourse constrained than the process in Spanish and Italian. For example, in Chinese, the empty subject can only refer to the subject of the preceding sentence 
\cite[see][]{liuModularTheoryRadical2014}. As 
a means of comparison, in surveying three Universal Dependencies datasets,\footnote{Chinese GSD, Italian ISDT, and Spanish AnCora.} 8\% of nsubj (or nsubj:pass) relations were 
pronouns for Chinese, while only 2\% and 3\% were pronouns in Spanish and Italian 
respectively. English lies on the opposite end of the continuum, 
requiring overt pronouns in the absence of other nominals 
(cf.\ \textit{He likes NLP} and *\textit{Likes NLP}). 

Therefore, it's possible that the presence of competing constraints in Spanish and Italian obscured the underlying IC knowledge: 
one constraint preferring pronouns which referred to the subject or object and the other constraint penalizing 
overt pronouns in subject positions (i.e.\ the target position masked in our experiments). In the following sections, 
we removed or otherwise demoted the 
dominance of each model's pro-drop constraint
for Spanish and Italian, and introduced or promoted a pro-drop like constraint 
in English and Chinese. We found
that the degree of IC bias in model behavior could be controlled by 
the presence, or absence, of a competing 
pro-drop constraint. 

\subsection{Methodology}

We constructed two classes of dataset to fine-tune the models on. The first aimed 
to demote the pro-drop constraint in Spanish and Italian. The second aimed to inject 
a pro-drop constraint into English and Chinese. For both we relied on 
Universal Dependencies datasets. For Spanish, we used the AnCora Spanish newswire corpus \cite{ancora}, for Italian we used ISDT \cite{bosco-etal-2013-converting} and VIT \cite{delmonteVITVeniceItalian2007}, 
for English we used the English Web Treebank \cite{silveira14gold}, and 
for Chinese, we used the Traditional Chinese Universal Dependencies Treebank annotated 
by Google (GSD) and the Chinese Parallel Universal Dependencies (PUD) corpus 
from the 2017 CoNLL shared task \cite{zeman-etal-2017-conll}.

\begin{figure}[t!] 
    \includegraphics[width=\linewidth]{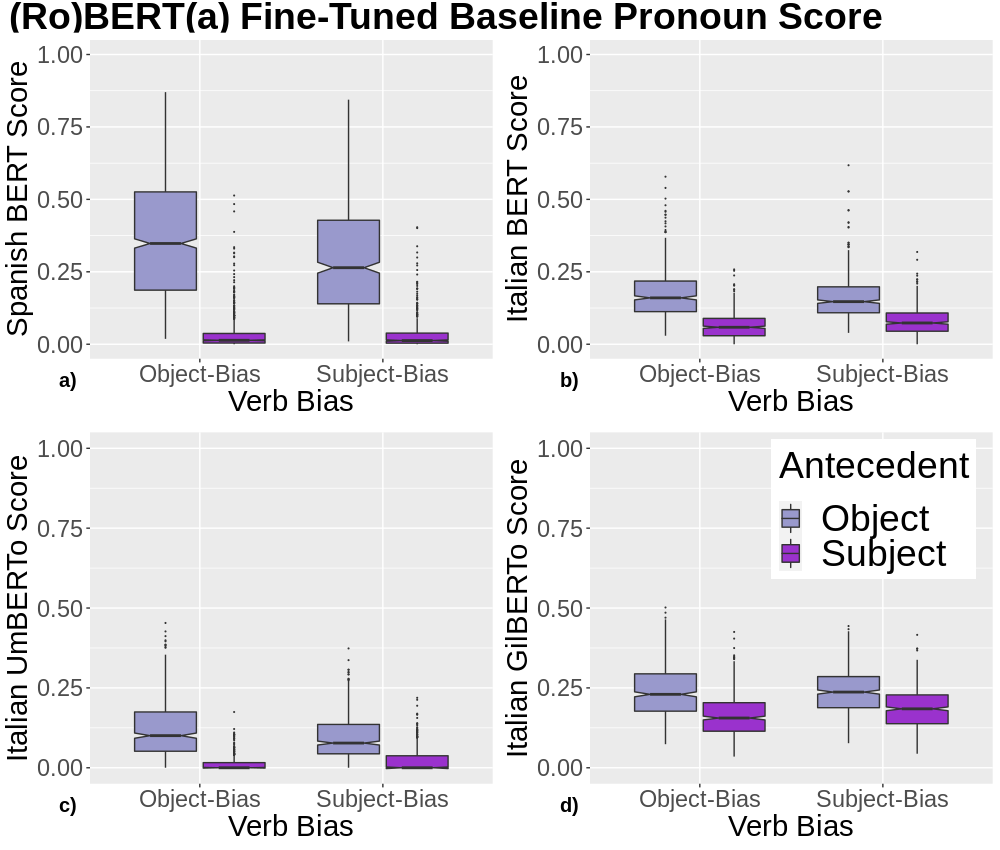}
    \caption{After fine-tuning on baseline data (i.e.\ pro-drop sentences), model scores for \textbf{a)} Spanish BERT (BETO), \textbf{b)} Italian BERT, \textbf{c)} UmBERTo, 
    and \textbf{d)} GilBERTo at the pronoun grouped by antecedent; stimuli
    derived from \citet{goikoetxeaNormativeStudyImplicit2008} and \citet{mannettiInterpersonalVerbsImplicit1991a}}
    \label{fig:es_it_pronoun_base}
\end{figure}

For demoting pro drop, we found finite (i.e.\ inflected) verbs that
did not have a subject relation in the corpora.\footnote{In particular, verbs that lacked any nsubj, nsubj:pass, expl, expl:impers, or expl:pass dependents} We then added a pronoun, matching 
the person and number information given on the verb, alternating the gender. For Italian, 
this amounted to a dataset of 3798 sentences with a total of 4608 pronouns (2,284 he or she) 
added. For parity with Italian, we restricted Spanish to a dataset of the first 4000 sentences, which had 5,559 pronouns (3,573 he or she) added. For the addition of a pro-drop constraint
in English and Chinese, we found and removed pronouns that 
bore a subject relation to a verb. This amounted to 935 modified sentences and 1083 removed 
pronouns (774 he or she) in Chinese and 4000 modified sentences and 5984 removed pronouns 
(2188 he or she) in English.\footnote{A fuller breakdown of the fine-tuning data is given in 
Appendix \ref{sec:fine_tune} with the full training and evaluation data given on 
our Github. We restricted English to the first 4000 sentences for parity with Italian/Spanish. Using the full set of sentences resulted in qualitatively the same pattern. We used the maximum
number of sentences we could take from Chinese UD.}

For each language, 500 unmodified sentences were used for validation, and unchanged versions of 
all the sentences were kept and used to fine-tune the models as a 
baseline to ensure that there was nothing about the data themselves that changed the 
IC-bias of the models. Moreover, the fine-tuning data was filtered to ensure that no verbs 
evaluated in our test data were included. Fine-tuning proceeded using HuggingFace's API.
Each model was fine-tuned with a masked language modeling objective for 3 epochs with a 
learning rate of 5e-5, following the fine-tuning details in \cite{devlinBERTPretrainingDeep2019}.\footnote{We provide a Colab script for reproducing 
all fine-tuned models on our Github.} 

\subsection{Demoting Pro Drop: Spanish and Italian}

As a baseline, we fine-tuned the Spanish and Italian models on unmodified versions 
of all the data we used for demoting pro drop. The baseline results are given in 
Figure \ref{fig:es_it_pronoun_base}. We found the same qualitative effects
detailed in Section \ref{sec:es_it_og}, confirming that the data used for fine-tuning on their own
did not produce model behavior in line with an IC bias. 

\begin{figure}[t!] 
    \includegraphics[width=\linewidth]{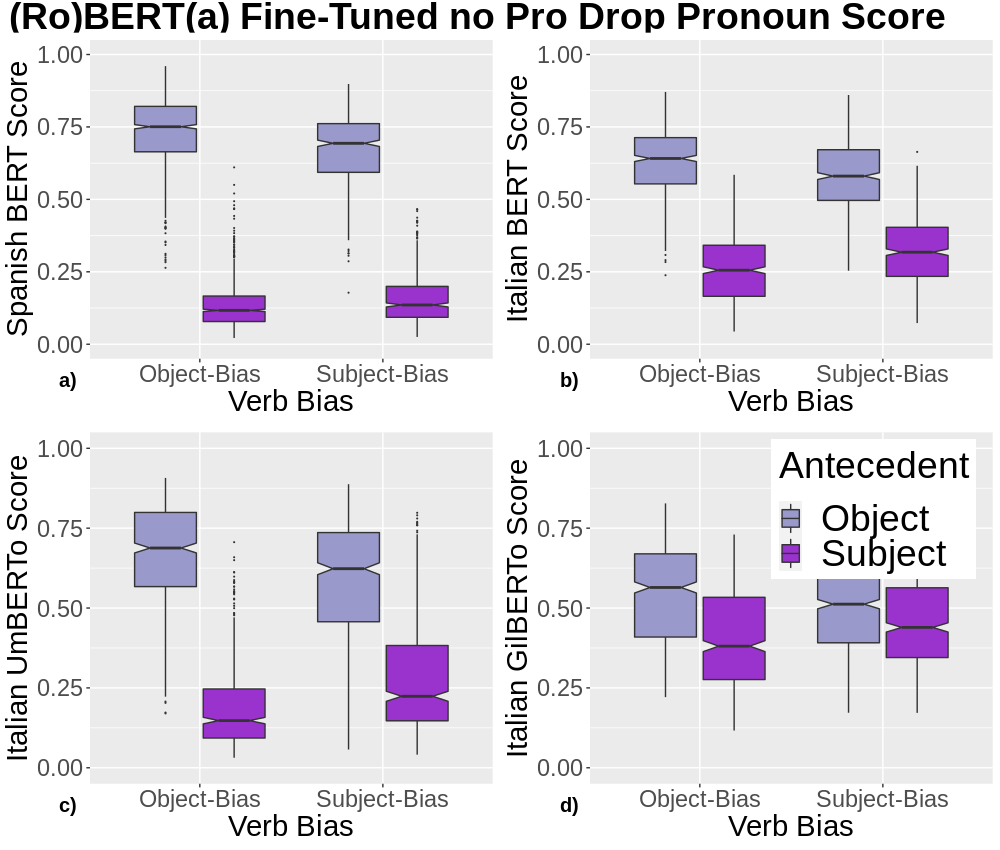}
    \caption{After fine-tuning on sentences removing pro drop (i.e.\ adding a subject pronoun), model scores for \textbf{a)} Spanish BERT (BETO), \textbf{b)} Italian BERT, \textbf{c)} UmBERTo, 
    and \textbf{d)} GilBERTo at the pronoun grouped by antecedent; stimuli
    derived from \citet{goikoetxeaNormativeStudyImplicit2008} and \citet{mannettiInterpersonalVerbsImplicit1991a}}
    \label{fig:es_it_pronoun_pro}
\end{figure}

We turn now to our main experimental manipulation: fine-tuning the Spanish and Italian 
models on sentences that exhibit the opposite of a pro-drop effect. It is worth repeating
that the fine-tuning data shared no verbs 
or sentence frames with our test data. 
The results are 
given in Figure \ref{fig:es_it_pronoun_pro}. Strikingly, an object-biased IC effect 
(pronouns referring to the object were more likely after object-biased IC verbs than 
subject-biased IC verbs) was observed for Italian BERT and GilBERTo despite no 
such effect being observed in the base models. Moreover, both models showed a more than doubled 
subject-biased 
IC verb effect. 
UmBERTo also showed increased IC effects, as 
compared to the base models. Similarly for Spanish, a subject-biased IC verb effect materialized for BETO when no corresponding
effect was observed with the base model. The object-biased IC verb effect remained similar
to what was reported in Section \ref{sec:es_it_og}. For RuPERTa, which showed no IC knowledge
in the initial investigation, no IC knowledge surfaced after fine-tuning. We conclude that RuPERTa has no underlying knowledge of IC, though further work should 
investigate this claim. 

\begin{figure}[t] 
    \includegraphics[width=\linewidth]{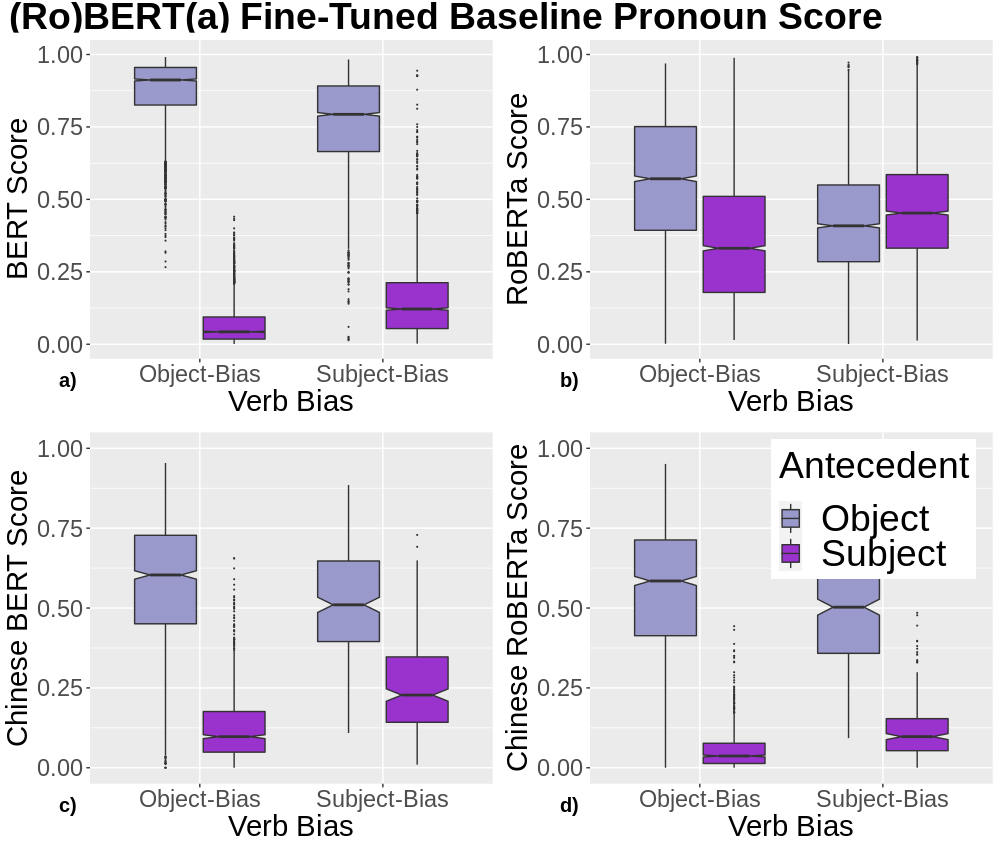}
    \caption{After fine-tuning on baseline data (i.e.\ without removing subject pronouns), model scores for \textbf{a)} BERT, \textbf{b)} RoBERTa, \textbf{c)} Chinese BERT, 
    and \textbf{d)} Chinese RoBERTa at the pronoun grouped by antecedent; stimuli
    derived from \citet{ferstlImplicitCausalityBias2011} and \citet{hartshorneAreImplicitCausality2013}}
    \label{fig:en_zh_pronoun_base}
\end{figure}

Taken together these results indicate that simply fine-tuning on a small number of sentences 
can re-rank the linguistic constraints influencing model behavior and
uncover other linguistic knowledge (in our 
case an underlying IC-bias). That is, model behavior can hide linguistic knowledge not just because of non-linguistic heuristics, but also due to over-zealously learning one isolated aspect of linguistic structure at the expense of another.

\subsection{Promoting Pro Drop: English and Chinese}

Next, we fine-tune a pro-drop constraint into models of English and Chinese. Recall 
that both models showed an IC effect, for both object-biased and subject-biased IC verbs. 
Moreover, both languages lack the pro-drop process found in Spanish and Italian (though 
Chinese allows null arguments). 

As with Spanish and Italian, we fine-tuned the English and Chinese models on 
unmodified versions of the training sentences
as a baseline (i.e.\ the sentences kept their pronouns) with the results given in Figure 
\ref{fig:en_zh_pronoun_base}. There was no qualitative difference
from the IC effects noted in Section \ref{sec:en_zh_og}. That is, for both English and 
Chinese, pronouns referring to the object were more likely after object-biased 
IC verbs than after subject-biased IC verbs, and conversely pronouns referring to the 
subject were more likely after subject-biased than object-biased IC verbs.

\begin{figure}[t] 
    \includegraphics[width=\linewidth]{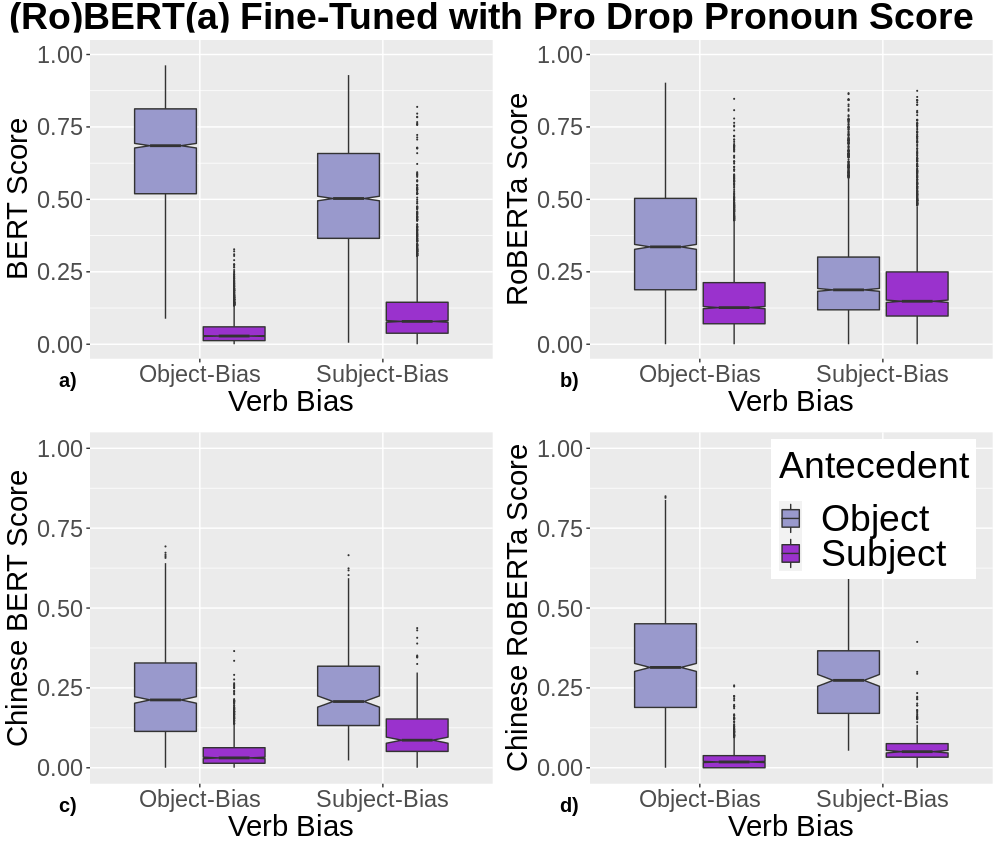}
    \caption{After fine-tuning on sentences with pro drop (i.e.\ no subject pronouns), model scores for \textbf{a)} BERT, \textbf{b)} RoBERTa, \textbf{c)} Chinese BERT, 
    and \textbf{d)} Chinese RoBERTa at the pronoun grouped by antecedent; stimuli
    derived from \citet{ferstlImplicitCausalityBias2011} and \citet{hartshorneAreImplicitCausality2013}}
    \label{fig:en_zh_pronoun_pro}
\end{figure}

The results after fine-tuning the models on data mimicking a Spanish and Italian 
like pro-drop process (i.e.\ no pronouns in subject position) are given in 
Figure \ref{fig:en_zh_pronoun_pro} and 
detailed in Appendix \ref{sec:full_results}.
Despite fine-tuning on only 0.0004\% and 
0.003\% of the data RoBERTa and BERT were trained on, respectively,  
the IC effects observed in Section \ref{sec:en_zh_og} were severely 
diminished in English. However, the 
subject-biased IC verb effect remained 
robust in both models. 
For Chinese BERT, the subject-biased IC verb 
effect in the base model was lost and the object-biased IC verb effect was 
reduced. The subject-biased IC verb effect was similarly 
attenuated in Chinese RoBERTa. However, the object-biased IC verb effect remained. 

For both languages, exposure to relatively little pro-drop data weakened the IC effect in behavior and even removed it in the case of subject-biased IC verbs in Chinese BERT.
This result strengthens our claim that competition between learned linguistic constraints can obscure underlying linguistic knowledge in model behavior.

\section{Discussion}

The present study investigated the ability of RoBERTa and BERT models 
to demonstrate knowledge of implicit causality across four languages (recall the contrast between 
\textit{Lavender frightened Kate} and \textit{Lavender admired Kate} in \ref{ic_exp}).
Contrary to humans, who show 
consistent subject and object-biased IC verb preferences across languages \cite[see][]{hartshorneAreImplicitCausality2013}, 
BERT and RoBERTa models of 
Spanish and Italian failed to demonstrate the full IC bias found in 
English and Chinese BERT and RoBERTa models \cite[with our English results supporting prior work on IC bias in neural models and extending it to non-autoregressive models;][]{upadhyePredictingReferenceWhat2020,davis-van-schijndel-2020-discourse}. Following standard behavioral probing 
\cite[e.g.,][]{linzenAssessingAbilityLSTMs2016}, 
this mismatch could be interpreted as evidence of 
differences in linguistic knowledge across languages. That is, model behavior in Spanish and Italian was 
inconsistent with predictions from the psycholinguistic IC literature, suggesting that these models lack knowledge of implicit causality. However, we found that to be an incorrect inference; the models \emph{did} have underlying knowledge of IC. 

Other linguistic processes influence pronouns in Spanish and Italian, and we showed that competition between multiple distinct constraints affects model behavior. 
One constraint (pro drop) decreases the probability of overt pronouns in subject position, while the other (IC) increases the probability of pronouns that refer to particular antecedents (subject-biased verbs like \textit{frightened} favoring 
subjects and object-biased verbs like \textit{admired} favoring objects). Models of Spanish and 
Italian, then, must learn not only these two constraints, but also their ranking (i.e.\ should 
the model generate a pronoun as IC dictates, or generate no pronoun in line with pro drop). 
By fine-tuning the models on data contrary to pro drop (i.e.\ with overt pronouns in subject 
position), we uncovered otherwise hidden IC knowledge. Moreover, we found that fine-tuning a pro-drop constraint into English and Chinese greatly diminished IC's influence
on model behavior (with as little as 0.0004\% of a models original training data). 

Taken together, we conclude that there are two ways of understanding mismatches between model linguistic behavior and human linguistic behavior. Either a model fails to learn the necessary linguistic constraint, or it 
succeeds in learning the constraint but fails to learn the correct interaction with other constraints. Existing literature points 
to a number of reasons a model may be unable to learn a linguistic representation, 
including the inability to learn mappings between form and meaning and the lack of embodiment \cite[e.g.,][]{benderClimbingNLUMeaning2020,biskExperienceGroundsLanguage2020a}.
We suggest that researchers should re-conceptualize linguistic inference on the part of neural models as inference of constraints and constraint ranking %
in order to better understand model behavior. We believe such framing will open additional connections with linguistic theory and psycholinguistics. %
Minimally, we believe targeted fine-tuning for 
constraint re-ranking may provide a general method both to understand
what linguistic knowledge these models possess and to aid in making their linguistic behavior more human-like.

\section{Conclusion and Future Work}

The present study provided evidence that model behavior can be meaningfully described, and 
understood, with reference to competing constraints. We believe that this is a 
potentially fruitful way of reasoning about model linguistic knowledge. 
Possible future directions include pairing our behavioral analyses with representational 
probing in order to more explicitly link model representations and model behavior \cite[e.g.,][]{ettingerProbingSemanticEvidence2016,hewittDesigningInterpretingProbes2019} or exploring constraint competition in
different models, like GPT-2 which has received considerable attention for its apparent 
linguistic behavior \cite[e.g.,][]{huSystematicAssessmentSyntactic2020} and its ability to predict 
neural responses \cite[e.g.,][]{schrimpfArtificialNeuralNetworks2020b}.

\section*{Acknowledgments}

We would like to thank members of the C.Psyd Lab, the Cornell NLP group, and the Stanford NLP Group, who gave valuable feedback on earlier forms of this work. Thanks also to the anonymous reviewers
whose comments improved the paper.

\bibliographystyle{acl_natbib}
\bibliography{acl2021}

\appendix

\section{Additional Fine-tuning Training Information}\label{sec:fine_tune}

The full breakdown of pronouns added or removed in the fine-tuning 
training data are detailed below. English can be found in Table \ref{tab:en_pronouns}, 
Chinese can be found in Table \ref{tab:zh_pronouns}, Spanish can be found 
in Table \ref{tab:es_pronouns}, and Italian can be found in Table \ref{tab:it_pronouns}. 

\begin{table}[h]
    \centering
    \begin{tabular}{c|c|c|c}
         & SG & PL & NA \\
         1 & 1927 & 617 & -\\
         2 & - & - & 1252 \\
         3 & 1548 & 640 & - \\
    \end{tabular}
    \caption{Breakdown of pronouns removed for English fine-tuning data. Pronoun person and number were determined by annotations in UD data, with NA being pronouns unmarked 
    for number. There were a total of 4000 sentences comprised of 66929 tokens in the training set. }
    \label{tab:en_pronouns}
\end{table}

\begin{table}[h]
    \centering
    \begin{tabular}{c|c|c|c}
         & SG & PL & NA \\
         1 & - & 56 & 66\\
         2 & - & 2 & 21 \\
         3 & - & 164 & 774 \\
    \end{tabular}
    \caption{Breakdown of pronouns removed for Chinese fine-tuning data. Pronoun person and number were determined by annotations in UD data, with NA being pronouns unmarked 
    for number. There were a total of 935 sentences comprised of 108949 characters in the training set. }
    \label{tab:zh_pronouns}
\end{table}

\begin{table}[h]
    \centering
    \begin{tabular}{c|c|c|c}
         & SG & PL & NA \\
         1 & 519 & 417 & -\\
         2 & 99 & 7 & - \\
         3 & 3574 & 944 & - \\
    \end{tabular}
    \caption{Breakdown of pronouns added for Spanish fine-tuning data. Pronoun person and number were determined by annotations in UD data, with NA being pronouns unmarked 
    for number. There were a total of 4000 sentences comprised of 5559 tokens in the training set. }
    \label{tab:es_pronouns}
\end{table}

\begin{table}[h]
    \centering
    \begin{tabular}{c|c|c|c}
         & SG & PL & NA \\
         1 & 654 & 417 & -\\
         2 & 399 & 94 & - \\
         3 & 2284 & 679 & - \\
    \end{tabular}
    \caption{Breakdown of pronouns added for Italian fine-tuning data. Pronoun person and number were determined by annotations in UD data, with NA being pronouns unmarked 
    for number. There were a total of 3798 sentences comprised of 4608 tokens in the training set. }
    \label{tab:it_pronouns}
\end{table}

\newpage

\section{Expanded Results (including mBERT)}\label{sec:full_results}

The full details of the pairwise \textit{t}-tests conducted for the present 
study are given below (including the results for mBERT). The results for 
English models are in Table \ref{tab:en_results}, for Chinese models 
Table \ref{tab:zh_results}, for Spanish models Table \ref{tab:es_results}, 
and Italian models Table \ref{tab:it_results}.

\begin{table*}[]
\resizebox{\linewidth}{!}{%
    \begin{tabular}{l|l|l|l|l|l|l|l|l}
         model &  O-O $\mu$ &  O-S $\mu$ & CI & p & S-O $\mu$ & S-S $\mu$ & CI & p  \\
         \hline
         BERT & 0.72 & 0.52 & [0.19,0.21] & $< 2.2e^{-16}$ & 0.13 & 0.26 & [0.12,0.13] & $<2.2e^{-16}$\\
         BERT\_BASE & 0.75 & 0.52 & [0.11,0.13] & $< 2.2e^{-16}$ & 0.06 & 0.15 & [0.08,0.09] & $<2.2e^{-16}$\\
         BERT\_PRO & 0.51 & 0.52 & [0.14,0.15] & $< 2.2e^{-16}$ & 0.04 & 0.11 & [0.06,0.07] & $<2.2e^{-16}$\\
         RoBERTa & 0.57 & 0.41 & [0.15,0.17] & $< 2.2e^{-16}$ & 0.31 & 0.43 & [0.11,0.13] & $<2.2e^{-16}$\\
         RoBERTa\_BASE & 0.58 & 0.45 & [0.11,0.13] & $< 2.2e^{-16}$ & 0.31 & 0.37 & [0.07,0.08] & $<2.2e^{-16}$\\
         
         RoBERTa\_PRO & 0.35 & 0.23 & [0.11,0.13] & $< 2.2e^{-16}$ & 0.16 & 0.19 & [0.03,0.04] & $<2.2e^{-16}$\\
         
         mBERT & 0.58 & 0.59 & [-0.003,-0.01] & 0.001 & 0.29 & 0.28 & [-0.002,-0.01] & 0.0002\\
    \end{tabular}}
    \caption{Results from pairwise \textit{t}-tests for English across the investigated models. O-O refers to object antecedent after object-biased IC verb and O-S to object antecedent after subject-biased IC verb (similarly for subject antecedents S-O and S-S). CI is 95\% confidence intervals (where positive is an IC effect). BERT\_BASE and BERT\_PRO refer to models fine-tuned on baseline data and data with a pro-drop process respectively. }
    \label{tab:en_results}
\end{table*}

\begin{table*}[]
\resizebox{\linewidth}{!}{%
    \begin{tabular}{l|l|l|l|l|l|l|l|l}
         model &  O-O $\mu$ &  O-S $\mu$ & CI & p & S-O $\mu$ & S-S $\mu$ & CI & p  \\
         \hline
         BERT & 0.41 & 0.39 & [0.003,0.05] & 0.00003 & 0.11 & 0.22 & [0.09,0.12] & $<2.2e^{-16}$\\
         BERT\_BASE & 0.53 & 0.47 & [0.03,0.08] & $2.2e^{-6}$ & 0.12 & 0.25 & [0.11,0.14] & $<2.2e^{-16}$\\
         BERT\_PRO & 0.23 & 0.23 & [-0.02,0.02] & 0.94 & 0.04 & 0.11 & [0.05,0.07] & $<2.2e^{-16}$\\
         RoBERTa & 0.40 & 0.33 & [0.04,0.08] & $1.16e^{-9}$ & 0.06 & 0.12 & [0.04,0.06] & $<2.2e^{-16}$\\
         RoBERTa\_BASE & 0.52 & 0.46 & [0.04,0.08] & $8.4e^{-7}$ & 0.05 & 0.11 & [0.05,0.07] & $<2.2e^{-16}$\\
         
         RoBERTa\_PRO & 0.32 & 0.29 & [0.002,0.06] & $7e^{-6}$ & 0.03 & 0.06 & [0.02,0.04] & $<2.2e^{-16}$\\
         
         mBERT & 0.08 & 0.07 & [0.01,0.03] & $2e^{-6}$ & 0.08 & 0.06 & [-0.009,-0.002] & $1.3e^{-5}$\\
    \end{tabular}}
    \caption{Results from pairwise \textit{t}-tests for Chinese across the investigated models. O-O refers to object antecedent after object-biased IC verb and O-S to object antecedent after subject-biased IC verb (similarly for subject antecedents S-O and S-S). CI is 95\% confidence intervals (where positive is an IC effect). BERT\_BASE and BERT\_PRO refer to models fine-tuned on baseline data and data with a pro-drop process respectively. }
    \label{tab:zh_results}
\end{table*}

\begin{table*}[]
\resizebox{\linewidth}{!}{%
    \begin{tabular}{l|l|l|l|l|l|l|l|l}
         model &  O-O $\mu$ &  O-S $\mu$ & CI & p & S-O $\mu$ & S-S $\mu$ & CI & p  \\
         \hline
         BERT & 0.53 & 0.46 & [0.04,0.09] & $1.4e^{-8}$ & 0.05 & 0.05 & [0.0007,0.01] & 0.03 \\
         BERT\_BASE & 0.37 & 0.30 & [0.05,0.08] & $8e^{-12}$ & 0.03 & 0.03 & [-0.004,0.007] & 0.61\\
         BERT\_PRO & 0.73 & 0.67 & [0.05,0.07] &  $<2.2e^{-16}$ & 0.16 & 0.13 & [0.01,0.03] & $1.2e^{-7}$\\
         RoBERTa & 0.09 & 0.10 & [-0.008,-0.01] & 0.03 & 0.06 & 0.06 & [0.0007,0.007] & 0.02 \\
         RoBERTa\_BASE & 0.06 & 0.06 & [-0.005,-0.002] & 0.0002 & 0.04 & 0.04 & [-0.0003,0.004] & 0.09\\
         
         RoBERTa\_PRO & 0.48 & 0.48 & [-0.03,0.01] & 0.42 & 0.29 & 0.30 & [-0.006,0.02] & 0.24\\
         
         mBERT & 0.12 & 0.11 & [0.001,0.01] & 0.02 & 0.02 & 0.02 & [-0.0002,-0.002] & 0.03\\
    \end{tabular}}
    \caption{Results from pairwise \textit{t}-tests for Spanish across the investigated models. O-O refers to object antecedent after object-biased IC verb and O-S to object antecedent after subject-biased IC verb (similarly for subject antecedents S-O and S-S). CI is 95\% confidence intervals (where positive is an IC effect). BERT\_BASE and BERT\_PRO refer to models fine-tuned on baseline data and data with a pro-drop process respectively. }
    \label{tab:es_results}
\end{table*}

\begin{table*}[]
\resizebox{\linewidth}{!}{%
    \begin{tabular}{l|l|l|l|l|l|l|l|l}
         model &  O-O $\mu$ &  O-S $\mu$ & CI & p & S-O $\mu$ & S-S $\mu$ & CI & p  \\
         \hline
         BERT & 0.21 & 0.19 & [0.005,0.03] & 0.004 & 0.09 & 0.11 & [0.01,0.03] & $1.3e^{-9}$ \\
         BERT\_BASE & 0.17 & 0.16 & [0.006,0.02] & 0.002 & 0.06 & 0.08 & [0.01,0.02] & $4e^{-6}$\\
         BERT\_PRO & 0.63 & 0.56 & [0.04,0.07] &  $1e^{-13}$ & 0.26 & 0.32 & [0.05,0.07] & $<2.2e^{-16}$\\
         UmBERTo& 0.06 & 0.05 & [0.01,0.02] & $4e^{-6}$ & 0.009 & 0.02 & [0.004,0.01] & $2e^{-9}$\\
         UmBERTo\_BASE & 0.12 & 0.09 & [0.02,0.04] & $3e^{-9}$ & 0.01 & 0.02 & [0.01,0.02] & $9e^{-12}$ \\
         UmBERTo\_PRO & 0.67 & 0.58 & [0.07,0.11] & $5e^{-16}$  & 0.19 & 0.28 & [0.07,0.11] & $<2.2e^{-16}$ \\
         
         GilBERTo& 0.26 & 0.25 & [-0.006,0.02] & 0.30 & 0.20 & 0.22 & [0.01,0.03] & 0.0002\\
         GilBERTo\_BASE & 0.24 & 0.24 & [-0.006,0.01] & 0.44 & 0.16 & 0.18 & [0.01,0.03] & $3e^{-7}$ \\
         GilBERTo\_PRO & 0.54 & 0.50 & [0.03,0.06] & $3e^{-7}$  & 0.40 & 0.45 & [0.04,0.07] & $3e^{-10}$ \\
         
         mBERT & 0.13 & 0.14 & [-0.004,-0.02] & 0.0003 & 0.12 & 0.13 & [0.003,0.02] & 0.003\\
    \end{tabular}}
    \caption{Results from pairwise \textit{t}-tests for Italian across the investigated models. O-O refers to object antecedent after object-biased IC verb and O-S to object antecedent after subject-biased IC verb (similarly for subject antecedents S-O and S-S). CI is 95\% confidence intervals (where positive is an IC effect). BERT\_BASE and BERT\_PRO refer to models fine-tuned on baseline data and data with a pro-drop process respectively. }
    \label{tab:it_results}
\end{table*}

\end{document}